\documentclass[letterpaper, 10 pt, conference]{ieeeconf}  
\usepackage{amssymb}
\usepackage{amsmath}
\usepackage{graphicx}
\usepackage{dcolumn}
\usepackage[font=footnotesize]{subcaption}
\usepackage[font=footnotesize]{caption}
\usepackage[hidelinks]{hyperref}
\usepackage{tikz}
\usetikzlibrary{shapes,arrows}
\usepackage{stfloats}
\usepackage{adjustbox}
\usepackage{siunitx}
\usepackage{pgfplots}
\usetikzlibrary{angles, quotes}
\usetikzlibrary{intersections}
\usetikzlibrary{fillbetween}
\usetikzlibrary{decorations.pathmorphing}
\usetikzlibrary{decorations.pathreplacing,calligraphy}
\usepackage{todonotes}
\usepackage{stfloats}
\usetikzlibrary{calc}
\usepackage{booktabs}
\usepackage{cleveref}
\usepackage{titlesec}

\DeclareSIUnit \vrms {\ensuremath{\mathrm{V_{rms}}}}
\DeclareSIUnit \vpp {\ensuremath{\mathrm{V_{pp}}}}

\usepackage[backend=biber, style=ieee, doi=false, url=false, eprint=false]{biblatex}
\tikzset{
    block/.style = {draw, rectangle, 
        minimum height=1cm, 
        minimum width=2cm},
    sum/.style = {draw, circle, node distance=2cm},
    input/.style = {coordinate},
    output/.style = {coordinate},
    arrow/.style={draw, -latex},
}

\IEEEoverridecommandlockouts                              

\title{\LARGE \bf
Self-Sensing Feedback Control of an Electrohydraulic Robotic Shoulder}

\author{Clemens C. Christoph$^{1}$, Amirhossein Kazemipour$^{1}$, Michel R. Vogt$^{1}$, Yu Zhang$^{1}$, Robert K. Katzschmann$^{*1}$
\thanks{$^{1}$ETH Zurich, Switzerland}
\thanks{\tt\small \{\href{mailto:cchristoph@ethz.ch}{cchristoph},\href{mailto:akazemi@ethz.ch}{akazemi},\href{mailto:micvogt@ethz.ch}{micvogt},\href{mailto:zhangyu@ethz.ch}{zhangyu},\href{mailto:rkk@ethz.ch}{rkk}\}
@ethz.ch}%
\thanks{$*$ Corresponding author: \href{mailto:rkk@ethz.ch}{\tt rkk@ethz.ch}}
}

\begin{document}
\maketitle
\thispagestyle{empty}
\pagestyle{empty}

\begin{abstract}
The human shoulder, with its glenohumeral joint, tendons, ligaments, and muscles, allows for the execution of complex tasks with precision and efficiency. However, current robotic shoulder designs lack the compliance and compactness inherent in their biological counterparts. A major limitation of these designs is their reliance on external sensors like rotary encoders, which restrict mechanical joint design and introduce bulk to the system.
To address this constraint, we present a bio-inspired antagonistic robotic shoulder with two degrees of freedom powered by self-sensing hydraulically amplified self-healing electrostatic actuators. Our artificial muscle design decouples the high-voltage electrostatic actuation from the pair of low-voltage self-sensing electrodes. This approach allows for proprioceptive feedback control of trajectories in the task space while eliminating the necessity for any additional sensors. We assess the platform's efficacy by comparing it to a feedback control based on position data provided by a motion capture system. The study demonstrates closed-loop controllable robotic manipulators based on an inherent self-sensing capability of electrohydraulic actuators. The proposed architecture can serve as a basis for complex musculoskeletal joint arrangements.

\end{abstract}

\section{Introduction}
The human shoulder stands out as an extraordinary biomechanical assembly, enabling complex movements to be carried out with precision and efficiency, thanks to its ball-and-socket joint, tendons, ligaments, and muscles. However, present-day robotic mechanisms that mimic this level of articulation in a single joint fall short in achieving the natural flexibility and streamlined design observed in biological systems. This shortfall primarily stems from their dependence on external sensors, such as rotary encoders, which limit the possibilities in mechanical joint construction and add unnecessary bulk. A solution is needed that imitates the human's musculoskeletal shoulder design, particularly the self-sensing muscles.

Spherical wrist mechanisms with multiple rotational degrees of freedom (DoF) are recognized for their excellent control capabilities; however, they are expensive and bulky, because each joint accommodates only one rotational DoF and thus requiring extra structures for each DoF~\cite{tri-motor-wire-encoder,tri-motos-encoder}. As the exploration into mechanisms, actuators, and integrated robotic joints with multiple degrees of freedom intensifies, it is evident that many are inspired by the principles governing the human shoulder. The promise of these mechanisms is clear: actuating multiple DoF within a single joint holds the potential for creating robotic systems that are more compact, cost-effective, and offer enhanced functionalities in comparison to their traditional single DoF counterparts. 

\begin{figure}[t!]
\centering   
\includegraphics[width=\linewidth]{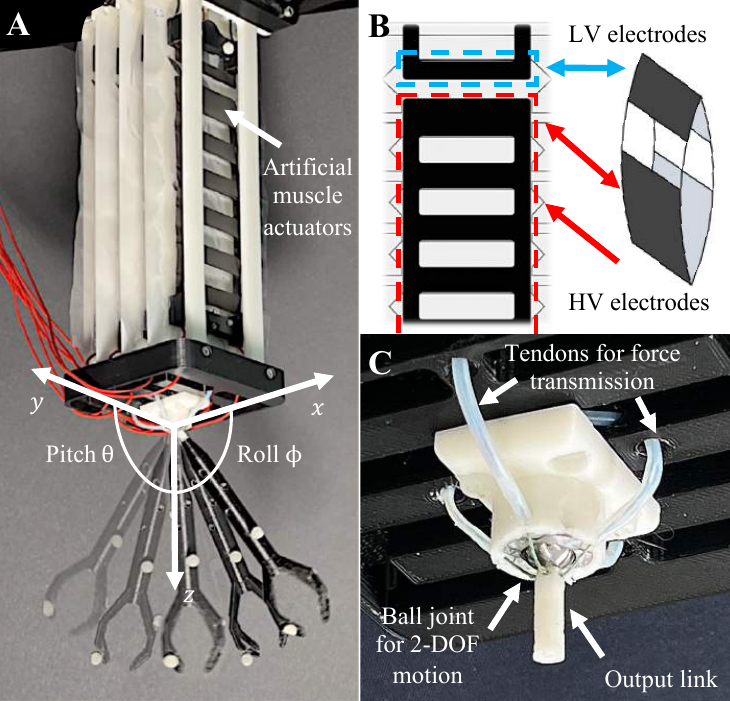}
\caption{(A) Bio-inspired robotic shoulder powered by two pairs of Peano-HASEL actuators with our self-sensing control capabilities. (B) Self-sensing HASEL design separating low voltage from high voltage electrodes for capacitive self-sensing. (C) Close-up of the ball-and-socket joint mechanism which allows for two-DoF motions: Tendons are attached to the output link of the ball joint, allowing the end effector to pivot up to \( \SI{80}{\degree} \) along the \( x \) and \( y \) axes.
}

\label{fig:main-pic}
\vspace{-20pt}
\end{figure}

A long-established method involves equipping a joint with friction wheels~\cite{friction-wheel} or omni-wheels~\cite{omni-wheel-1, omni-wheel-2, omni-wheel-3} driven by DC-motors for force transmission. Omni-wheels offer greater efficiency than friction wheels due to lower frictional losses. However, they tend to slip under high torques caused by substantial loads or rapid movement changes. Consequently, external sensors become necessary to achieve reliable positional feedback control. To address the issue of slippage, a gear-based active ball joint mechanism, ABENCIS~\cite{abenics}, has been introduced. This mechanism is capable of transferring high torques while achieving impressive control capabilities. However, the mechanism faces challenges with low backdrivability due to its high gearing ratio.
Alternative methods have sought inspiration from biological systems, employing linear actuators. For instance, the human-like shoulder complex~\cite{pam-shoulder} uses PAMs (pneumatic artificial muscles), while another design integrates linear piston actuators on a ball joint~\cite{linear-motors}. Yet, these systems do not achieve closed-loop control due to the absence of essential sensor data from these actuator types.

When it comes to controlling these systems in a closed-loop manner, one common challenge is their dependence on external sensors. Rotary encoders, for instance, are employed for actuating ball joints with motors such as friction or gear mechanisms, but their integration can increase the system's weight and costs. A different method involves incorporating sensors right into the ball joint to passively detect its rotation. A popular technique uses permanent magnets paired with hall sensors~\cite{hall-sensor-pm-inside-ball-joint, hall-sensor-pm-on-rod, hall-sensor-suggestion} , and magnetic field sensors~\cite{magnetic-field-based-sensor}. Another approach is optical sensing~\cite{optical-sensor-joints}. Although they have good sensing accuracy, those sensors pose design limitations; their size and critical positioning relative to magnets can make them unsuitable for intricate ball joint configurations. Other methods that involve direct contact with the ball joint~\cite{friction-contact} can result in undesirable frictional losses. Lastly, inertial measurement units (IMUs) can be appropriate for end effectors, but cannot always be integrated within a joint.

\begin{figure}
    \centering
    \includegraphics[width=\linewidth]{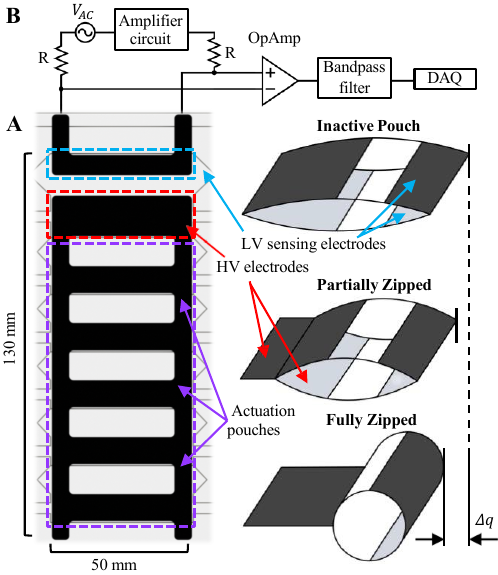}
    \caption{(A) HASEL design featuring five actuation pouches, each measuring \(w =\) \SI{50}{mm} by \(h = \) \SI{20}{mm}. Placed on top is the self-sensing pouch sized \(w =\) \SI{50}{mm} by \(h =\) \SI{30}{mm}. This sensing pouch detects the voltage difference across its low-voltage electrodes. When the pouch is inactive, the sensing electrodes are closely spaced, resulting in a high capacitance and a corresponding low voltage drop. When fully actuated at displacement \(\Delta q\), the capacitance of the low-voltage electrodes is at its lowest, resulting in the highest possible voltage drop as the pouch reaches its maximum displacement. (B) Electrical self-sensing circuit: A \SI{10}{\volt} peak-to-peak \SI{2}{kHz} sinusoidal signal is generated. The signal is then amplified by a non-inverting operational amplifier circuit  (OP07CD, Texas Instruments) with a gain of \(1:2\). The signal path consists of two resistors of equal value (\SI{1}{\mega\ohm}) as well as the sensing electrodes of the Peano-HASEL, all connected in series. An instrumentation amplifier (INA821, Texas Instruments) monitors the voltage drop over the sensing electrodes and passes the output through a band-pass filter (\(f_c = \) \SI{2}{\kHz}) to the data acquisition (DAQ) system. }
    \label{hasel-design}
    \vspace{-20pt}
\end{figure}

Drawing inspiration from the human shoulder, this work contributes a bio-inspired robotic shoulder with self-sensing control capabilities (as seen in \Cref{fig:main-pic}). 
The cornerstone of our approach is a ball-and-socket joint actuated by antagonistically paired HASEL actuators~\cite{hasel-muscles-introduction}. Self-sensing control of HASEL actuators has been demonstrated before~\cite{ly_self_sensing_control_2021, system-id-self-sensing-feedback}, as well as control methods using embedded magnetic or capacitive strain sensors~\cite{magnetic-sensor-hasel, control-capacitive-sensor}. However, the use of self-sensing has been limited to one-DoF robotic arms or test rig experiments. Also, these setups do not have an antagonistic configuration and, as a result, depend on gravitational forces. 

To the best of our knowledge, there exists no design of a spherical robotic shoulder joint powered by HASEL actuators. Moreover, no multi-DoF system has been reported that achieves self-sensing closed-loop control. We contribute the following features with our integrated system:

\begin{itemize}
\item A compliant design backed by soft artificial muscle actuators with a new self-sensing method.
\item Elimination of external sensors, thanks to our reliable  self-sensing capabilities of HASEL actuators. This freedom ensures that the sensor integration cannot affect our compact joint design (radius \SI{4}{mm}).
\item Use of tendons for torque transmission, which minimizes frictional losses. This design choice enables quick changes in movement while ensuring stable actuation at \SI{3}{Hz}.
\item Our system can estimate and control the end-effector position in the task space.
\end{itemize}

\section{System Design}
\label{system-design}
In this section, we describe the design aspects of the system, including the mechanical, electrical, and software domains. The focus will be on the actuator design, ball joint design, and controller implementation. All components, with the exception of the joint's ball and bearing, are tailored for rapid prototyping, ensuring a lightweight design of \SI{272}{g}\footnote{Weight excludes power electronics and computing devices.} and cost-effective fabrication. We begin by defining specific components of the system:

\subsection{Terminology}
In the Cartesian \textit{task space}, the end effector's position of the manipulator (length \(l_m\)) is defined as \(\mathbf{x} = \mathbf{R}\mathbf{x_0} \) where \(\mathbf{x_0} = (0,0,-l_m)\) is the initial position (pointing downwards) and \(\mathbf{R} \in \mathbb{R}^{3\times3}\) is the rotation matrix. We define the origin  \( \mathbf{O} = (0,0,0)\) as the fixed center of the ball joint. In the \textit{joint space}, the roll angle ($\phi$), pitch angle ($\theta$), and yaw angle ($\psi$) define the position of the manipulator. The actuation system can only apply forces in $x$ and $y$ axes and thus $\psi = 0$. Components in the \textit{tendon space} such as HASELs ($H_i$), displacement ($q_i$), and tendons ($T_i$) are denoted with subscripts $i = \{ \phi_1, \phi_2, \theta_1, \theta_2\}$ depending on the direction and element of an antagonistic pair to which they correspond.

\subsection{Design of Actuator and Self-Sensing}
The Peano-HASEL actuators utilized in this study are composed of multiple pouches to enhance the linear displacement and, consequently, the range of motion of the manipulator. To maintain a compact system, we restricted the number of pouches to six. Out of these six, five pouches are designated as actuation pouches, serving solely for actuation purposes. The sixth pouch is a customized design, distinguished by the addition of a second electrode placed at the edge of the pouch. This configuration is intended to enable capacitive self-sensing capabilities (see \Cref{hasel-design}). Over these sensing electrodes, a low voltage sinusoidal signal \( V_{\text{AC}} \) (frequency \(\omega\))
 is applied.  An off-the-shelf instrumentation amplifier is used to monitor the voltage drop \(V^H\) over the sensing electrodes at all times. When electrically modeling the sensing electrodes as a variable capacitor with capacitance \(C_H\), the voltage drop \(V^H\) will change as the pouch deforms. This is caused by the behavior of the first-order low pass filter that the variable capacitor forms with any resistances \(R\) placed in series to the sensing electrodes:
\begin{equation}
    \frac{V^{H}}{V_\text{AC}} = \frac{1}{1+j \omega R C_H}.
\end{equation}
As such, the measured change in \(V^H\) can be used to directly estimate the displacement of the pouch as the measured increase in voltage gain is directly proportional to the displacement \(q\) of the HASEL actuator. This method allows for the measurement of both active displacement and passive displacement under external loads. For data processing, we consider the RMS value of \(V^H\). Henceforth, we will denote the RMS voltage drop of HASEL \(i\) as \(V_i^H\). The pouch size was minimized for efficiency ~\cite{hasel-muscles-introduction} yet sufficiently large to ensure effective voltage measurement by the self-sensing electrode.

For our system, we used four of the described HASELs and we obtained the achievable strain at different loads on a test rig and measured the displacement with a laser sensor. All HASELs were filled with silicone oil at \SI{80}{\percent} of the volume of the cylinder in a fully actuated state.

\begin{table}[h]
\centering
\caption{HASEL Characteristics at \SI{5}{kV} step signal}
\begin{tabular}{ccccc}
\toprule
 & \multicolumn{2}{c}{14g Load} & \multicolumn{2}{c}{34g Load} \\
\cmidrule{2-5}
 & Strain & Displacement (mm) & Strain & Displacement (mm) \\
\midrule
$H_{\phi,1}$ & 0.065 & 8.45 & 0.0588 & 7.65 \\
$H_{\phi,2}$ & 0.0642 & 8.35 & 0.0575 & 7.47 \\
$H_{\theta,1}$ & 0.0648 & 8.42 & 0.0586 & 7.62 \\
$H_{\theta,2}$ & 0.0662 & 8.60 & 0.0601 & 7.81 \\
\bottomrule
\label{tab:hasel_geom}
\end{tabular}
\end{table}

\Cref{tab:hasel_geom} shows that the strain of the actuators varies due to fabrication uncertainties. The system's ability to lift higher payloads depends on the force strain characterization of the HASELs; that is, the strain decreases at higher loads, which is a typical trait of the force-strain characterization of HASELs~\cite{hasel-muscles-introduction}. To overcome this problem, we use a lead screw mechanism that can adjust the height of the HASELs and thus the slack on the tendons. This mechanism allows us to control the range of motion of the rotational axes. When the manipulator load is high, we reduce the slack and the range of motion, but the manipulator remains controllable.

\subsection{Ball Joint Design}
\label{ball-joint}

\begin{figure}
\centering
\includegraphics[width=\linewidth]{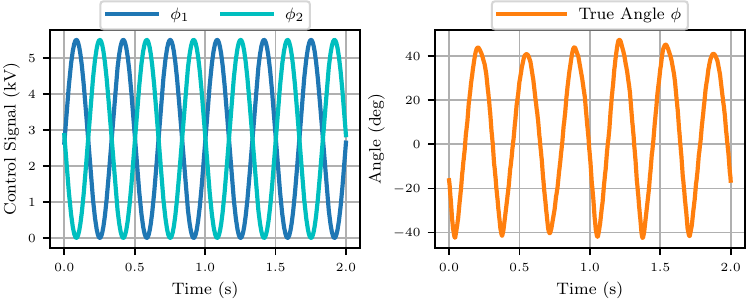}
\caption{Open-loop range for the \(\phi\) angle. A \SI{5.5}{\kV} \SI{3}{\Hz} sinusoidal signal is applied to \(H_{\phi,1}\) and \(H_{\phi,2}\) with a half-period phase shift.}
\label{fig:range-open-loop}
\vspace{-20pt}
\end{figure}

The achievable displacement of our HASEL actuators, which is approximately \SI{8}{mm}, dictates the design constraints for the system's ball joint. Given this constraint, we selected a spherical joint radius of \(r_s = \SI{4}{mm}\) to guarantee a broad range of motion for the end effector. To avoid friction between the joint and the tendons $r_t$ was chosen chosen sufficiently large.
A key consideration for ensuring smooth movement and achieving precise control is minimizing friction between the bearing and the sphere and between the sphere and the fixture. We used a steel ball with an aluminum bearing for our design, while the fixture was 3D-printed using a resin printer. Tendons are attached to the output link of the ball joint, allowing for the direct transmission of forces from the HASELs to the manipulator, thereby mitigating potential slippage. Furthermore, the tendon arrangement, as illustrated in \Cref{fig:subfig1}, was designed to optimize force transmission. Specifically, when the HASEL is fully contracted—such as when gravity pulls the output link downward requiring the actuator's maximum force output—the tendons are oriented perpendicular to the output link. This positioning enhances force application, thus increasing the range of motion. We validated the range of motion by subjecting both HASEL pairs to a \SI{5.5}{kV}, \SI{3}{Hz} sinusoidal signal, achieving a deflection of more than \SI{80}{\degree} (see \Cref{fig:range-open-loop}).

\begin{center}
\begin{figure}
    \centering
    \label{fig:ball-joint-fig}
\begin{adjustbox}{valign=t,minipage={.5\linewidth}}

    \begin{subfigure}{.9\linewidth}
\hfill
\begin{tikzpicture}
    \def\diameter{1.3}
    \def\thickness{1pt}
    \draw[line width=\thickness*2] (-\diameter*1.35,-\diameter*0.48) .. controls ({-\diameter*0.5}, {-1*\diameter}) ..({1.36*\diameter*cos(75)}, {-1.35*\diameter*sin(75)})
    node[midway, above left, inner sep=2pt, label={[label distance=-5pt]-115:$T_{\phi,2}$}] {};
    \draw[line width=\thickness*2] (+\diameter*1.35,-\diameter*0.48) -- ({1.36*\diameter*cos(60)}, {-1.45*\diameter*sin(60)})
    node[midway, above right, inner sep=2pt, label={[label distance=-5pt]-65: $T_{\phi,1}$}] {};
    
    \fill[gray!30] (-\diameter*1.5,+\diameter*0.5) rectangle (\diameter*1.5,+\diameter*1.2);
    \node[anchor=north west, font=\captionlabelfont\captionfont] at (\diameter*0.7,+\diameter*1.1) {Fixture};
    \fill[gray!30] (-\diameter*1.5,-\diameter*0.5) rectangle (-\diameter*1.1,+\diameter*0.5);
    \fill[gray!30] (+\diameter*1.5,-\diameter*0.5) rectangle (+\diameter*1.1,+\diameter*0.5);
    
    \fill[gray!50] (-\diameter*1.5,-\diameter*0.5) rectangle (\diameter*1.5,-\diameter*0.5 + \diameter*0.1);
    \fill[white] (0,0) circle (1.05*\diameter);
    \node[anchor=north,font=\captionlabelfont\captionfont] at (\diameter*1.4,-\diameter*0.1) {Bearing};
    \fill[gray!30] (0,0) -- ({2*\diameter*cos(75)}, {-2*\diameter*sin(75)}) arc (285:315:2*\diameter) -- cycle;
    \node[anchor=north west, font=\captionlabelfont\captionfont] at ({1.7*\diameter*cos(75)}, {-1.8*\diameter*sin(75)}) {Output Link};
    \fill[gray!50] (0,0) circle (\diameter);
    \draw[dashed,line width=\thickness/2] (0, 0) -- ({2*\diameter*cos(60)}, {-2*\diameter*sin(60)});
    \draw[dashed,line width=\thickness/2] (0,0) -- (0, -\diameter*2);

    \node at (-\diameter/25, \diameter/12) {$\odot$};
    \node at (0, \diameter/12) [above left] {$x$};
    \draw[->,line width=\thickness/2] (0,0) -- (-\diameter/2,0) node[left] {$y$};
    \draw[->,line width=\thickness/2] (0,0) -- (0,-\diameter/2) node[below left] {$z$};
    
     \draw[->,line width=\thickness/2] (0,0) -- ({\diameter*cos(45)},{\diameter*sin(45)});
    \node at ({0.4*\diameter*cos(45)}, {0.7*\diameter*sin(45)}) {$r_s$};
    \draw[dashed] ({1.35*\diameter*cos(285)}, {1.35*\diameter*sin(285)}) arc (285:315:1.3*\diameter);
    \draw[->,line width=\thickness/2] (0, 0) -- ({1.35*\diameter*cos(60)}, {-1.35*\diameter*sin(60)});
    \node at ({0.75*\diameter*cos(60)}, {-0.4*\diameter*sin(60)}) {$r_t$};
\end{tikzpicture}

\caption{Cross section schematic of the ball and socket joint in a deflected state where $H_{\phi,1}$ is contracted and $H_{\phi,2}$ is relaxed.}
        \label{fig:subfig1} 
    \end{subfigure}%
\end{adjustbox}%
\begin{adjustbox}{valign=t,minipage={.5\linewidth}}
    \begin{subfigure}{\linewidth}
\hfill
\begin{tikzpicture}
    \def\cdiameter{1.5}
    \draw[dashed, gray] (0,0) circle (\cdiameter cm);
    \draw [decorate, decoration={brace, amplitude=7pt}] (0,0) --  (-\cdiameter cm,0) node[midway, below=5pt] {\( q_t \)};
    \draw[dashed,black] (0.48*\cdiameter,0.28*\cdiameter) circle (0.2*\cdiameter cm);
    \draw[->] (-\cdiameter*1.2 cm,0) -- (\cdiameter*1.2 cm,0) node[right] {$x$};
    \draw[->] (0,-\cdiameter*1.2 cm) -- (0,\cdiameter*1.2 cm) node[above] {$y$};
    \draw (\cdiameter*0.48, 2pt) -- (\cdiameter*0.48, -2pt) node[anchor=north] {$x'$};
    \draw (2pt, \cdiameter*0.28) -- (-2pt, \cdiameter*0.28) node[anchor=east] {$y'$};
    \draw[dashed] (\cdiameter*0.48,0) -- (\cdiameter*0.48,\cdiameter*0.28);
    \draw[dashed] (0,\cdiameter*0.28) -- (\cdiameter*0.48,\cdiameter*0.28);
    \draw[blue] (\cdiameter cm,0) -- (0.48*\cdiameter,0.28*\cdiameter);
    \draw[blue] (-\cdiameter cm,0) -- (0.48*\cdiameter,0.28*\cdiameter);
    \draw[blue] (0,\cdiameter) -- (0.48*\cdiameter,0.28*\cdiameter);
    \draw[blue] (0,-\cdiameter) -- (0.48*\cdiameter,0.28*\cdiameter);
\end{tikzpicture}

\caption{Projection of the four tendons onto a plane to find an approximation of the HASEL displacement given a certain Cartesian position.}
        \label{fig:subfig2} 
    \end{subfigure}
\end{adjustbox}
\caption{Ball joint design and inverse kinematic model.}

\end{figure}
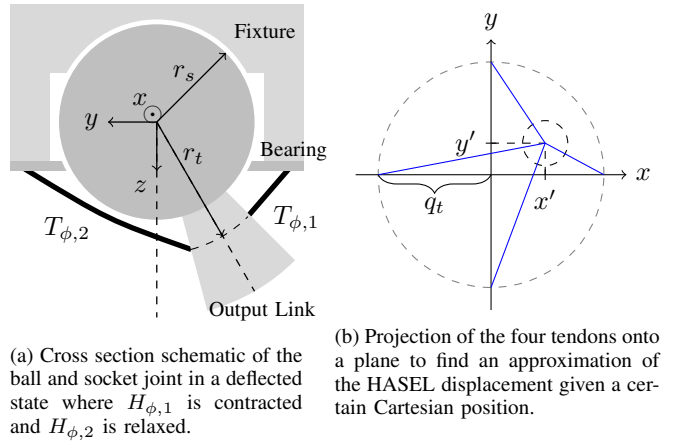
\end{center}

\vspace{-23pt}
\subsection{Control Architecture}

\begin{figure}
\centering
\includegraphics[width=\linewidth]{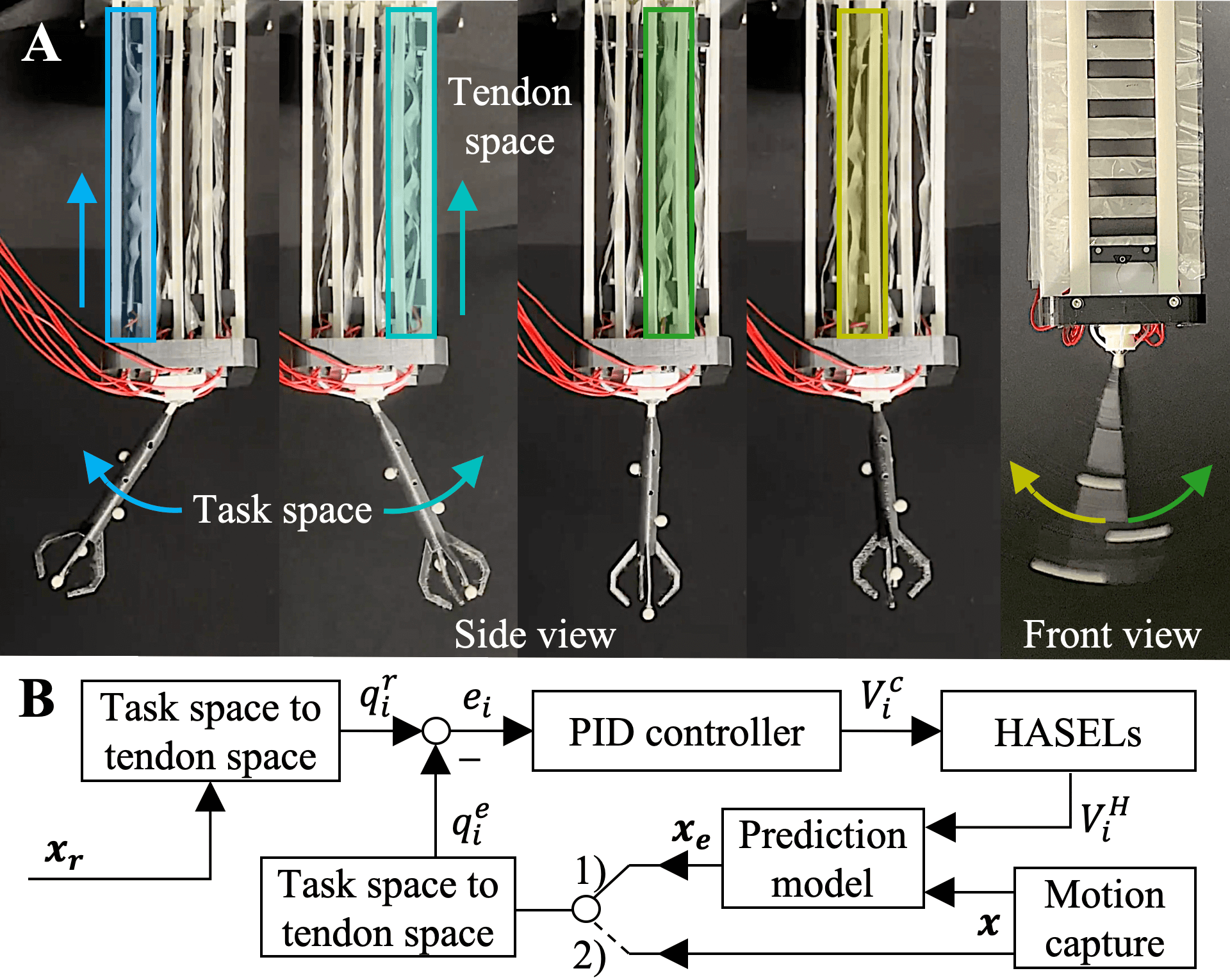}
\caption{(A) Working principle of the system: By contracting and relaxing the artificial muscle pairs, the manipulator can reach the desired pitch and roll angles. (B) Proposed PID feedback control architecture for \(i = \{ \phi_1, \phi_2, \theta_1, \theta_2\}\). The controller supports two configurations: 1) Self-sensing feedback control based on estimations \(\mathbf{x_e}\) without external sensors. 2) Feedback control using motion capture position data \(\mathbf{x}\) (providing ground truth data). This data serves as a benchmark while also training and validating the prediction model.}
\vspace{-20pt}
\label{fig:controller}
\end{figure}

As an initial step towards more advanced control, we use a PID controller for the HASELs. System identification from past studies indicates that HASELs adhere well to linear control models \cite{keplinger_HASELs}.
The control loop is depicted in \ref{fig:controller}B. The reference signal \(\mathbf{x_{r}} = (x_r,y_r,z_r) \), represents the desired end effector position in the task space, lying on a spherical surface with radius $l_{m}$. This signal is subsequently transformed into the tendon space to represent desired HASEL displacements, \({q_i^r}\). Upon receiving a feedback signal \({q_i^e}\), an error is generated and routed to the PID controller. The controller processes the inputs and commands the high voltage signals \({V_i^c}\) to actuate each HASEL individually. As the HASEL state alters, so does the self-sensing voltage signal \({V_i^H}\), which we feed to an estimation model that maps the measured voltages to an estimated Cartesian position \(\mathbf{x_{e}} = (x_e,y_e,z_e) \). To compare the feedback signal with the reference, we map the feedback to the tendon space \({q_i^e}\) and close the loop. The controller also supports feedback using motion capture position data \(\mathbf{x} = (x,y,z) \), providing ground truth. This serves as a benchmark while also training and validating the prediction model.

\subsection{Mapping task space to tendon space}

To control all four HASEL actuators, we need to establish the mapping \(  \mathcal{T}: \mathbb{R}^3 \to \mathbb{R}^4\), where
\begin{equation}
    \mathcal{T}(x,y,z) = (q_{\phi,1},q_{\phi,2},q_{\theta,1},q_{\theta,2}),
\end{equation} 
from task space to the tendon space.
For instance, to rotate the \(\phi\) angle, \(H_{\theta,1}\) and \(H_{\theta,2}\) need to relax to allow for the desired rotation. Deriving an accurate analytical model is challenging due to the tendons bending at specific angles and the three-dimensional nature of the problem. Moreover, potential errors from manufacturing and assembly processes can further complicate the task.
However, since we designed the controller to map both reference and feedback signals from the task space to the tendon space, a rough approximation is adequate, because minimizing the error in the tendon space will also minimize the error in the task space. For our approximation, we first find the attachment point of the tendons to the output link \( \mathbf{x_t} = \mathbf{R} \begin{bmatrix} 0 & 0 & -r_t \end{bmatrix}^T \)
where \( r_t = \SI{5}{mm} \) represents the radial distance from the origin to the attachment point, as depicted in \ref{fig:subfig1}. We then project \(\mathbf{x_t} \) as well as the tendons onto the \(x-y\) plane going through the origin (see \Cref{fig:subfig2}), resulting in \(\mathbf{x'}=(x',y',0)\).
Using geometry and defining \( q_t = \SI{3.75}{mm} \) as half of the maximum HASEL displacement, we can determine the approximate solution for \(\mathcal{T}\):
\begin{alignat}{2}
    q_{\phi,1} &= \sqrt{(q_t - x')^2 + y'^2}, &\quad
    q_{\phi,2} &= \sqrt{(q_t + x')^2 + y'^2} \notag, \\
    q_{\theta,1} &= \sqrt{x'^2 + (q_t-y')^2}, &\quad
    q_{\theta,2} &= \sqrt{x'^2 + (q_t+y')^2} \notag .
\end{alignat}

\subsection{Mapping  Self-Sensing Voltages to Displacement}
\begin{figure}[t]
    \label{relationship}
    \centering
    \includegraphics[width=\linewidth]{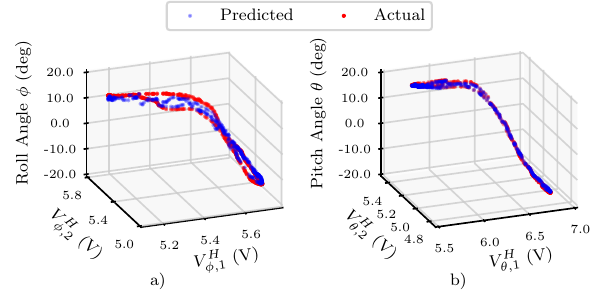}
    \caption{The \(\phi\) and \(\theta\) angles were subjected to a sinusoidal signal of \SI{5}{\kV} at \SI{1}{\Hz}. The predicted values were obtained using a third-order polynomial regression with a test size of \(0.2\). (a) Relationship between $V_{\phi,1}^H,V_{\phi,1}^H$ and $\phi$. (b) Relationship between $V_{\theta,1}^H,V_{\theta,1}^H$ and $\theta$.
    }
    \vspace{-10pt}
    \label{fig:relationship}
\end{figure}

The key element that allows us to use the self-sensing signal as feedback of the two-DoF antagonistic system is the mapping \(  \mathcal{V}: \mathbb{R}^4 \to \mathbb{R}^3\)
\begin{equation}
    \mathcal{V}(V_{\phi,1}^H,V_{\phi,2}^H,V_{\theta,1}^H,V_{\theta,2}^H) = (x_e,y_e,z_e).
    \label{eqn:v-mapping}
\end{equation} 
Given the slight variations that might be introduced in fabrication and potential non-homogeneity in zipping of actuator pouches, we chose to refrain from using first-principle methods to derive an analytical model for \(\mathcal{V}\).
Furthermore, instead of mapping each \(V_i^H\) separately to the tendon space, we directly relate the voltages to the estimated position in the task space. We adopted this approach for two primary reasons. First, the implemented controller needs to map both the reference and feedback from the task to the tendon space to ensure the accurate following of the trajectory within the task space. Second, determining the estimated end effector's cartesian position from four estimated tendon positions presents inherent challenges, requiring a separate optimization model.
We first map to the joint space, which more closely correlates with tendon positions, helping to linearize the prediction model. We actuated the $\phi$ and $\theta$ angles using an open-loop signal and analyzed the relationship between measured voltages and the actual angular deflection (see \Cref{fig:relationship}).
To train the model we use ground truth from the motion capture system. We acquire the rotational matrix \( \mathbf{R} = \left[ r_{11}, r_{12}, r_{13}; \ldots;\ldots, r_{33} \right] \) and compute the true Euler angles \( \phi = \operatorname{atan2}\left(r_{32}, r_{33}\right) \) and \( \theta = \operatorname{atan2}\left(-r_{31}, \sqrt{r_{32}^2 + r_{33}^2}\right) \).

\Cref{fig:relationship} shows that a third-order polynomial regression effectively captures the correlation.
The map \(f\) being learned by the third-order polynomial regression is:
\begin{equation}
    \underbrace{(\phi_e, \theta_e)}_{\text{output}} = f(\underbrace{V_{\phi,1}^H,V_{\phi,2}^H,V_{\theta,1}^H,V_{\theta,2}^H)}_{\text{input}}.
\end{equation}

After estimating rotation angles in the joint space, we derive the rotation matrix \(\mathbf{R}\), assuming a \(\psi = 0\), which defines the position of the manipulator yielding the mapping \(\mathcal{V}\) specified in equation \ref{eqn:v-mapping}.

\section{Results and Discussion}
\label{results-discussion}
In this section, we delve into the experimental setup, emphasizing the hardware and software employed. We then present and discuss the results of closed-loop control for task space (lemniscate and star-shaped) trajectories.

\subsection{Experimental Setup}
\begin{figure}
\centering
\includegraphics[width=\linewidth]{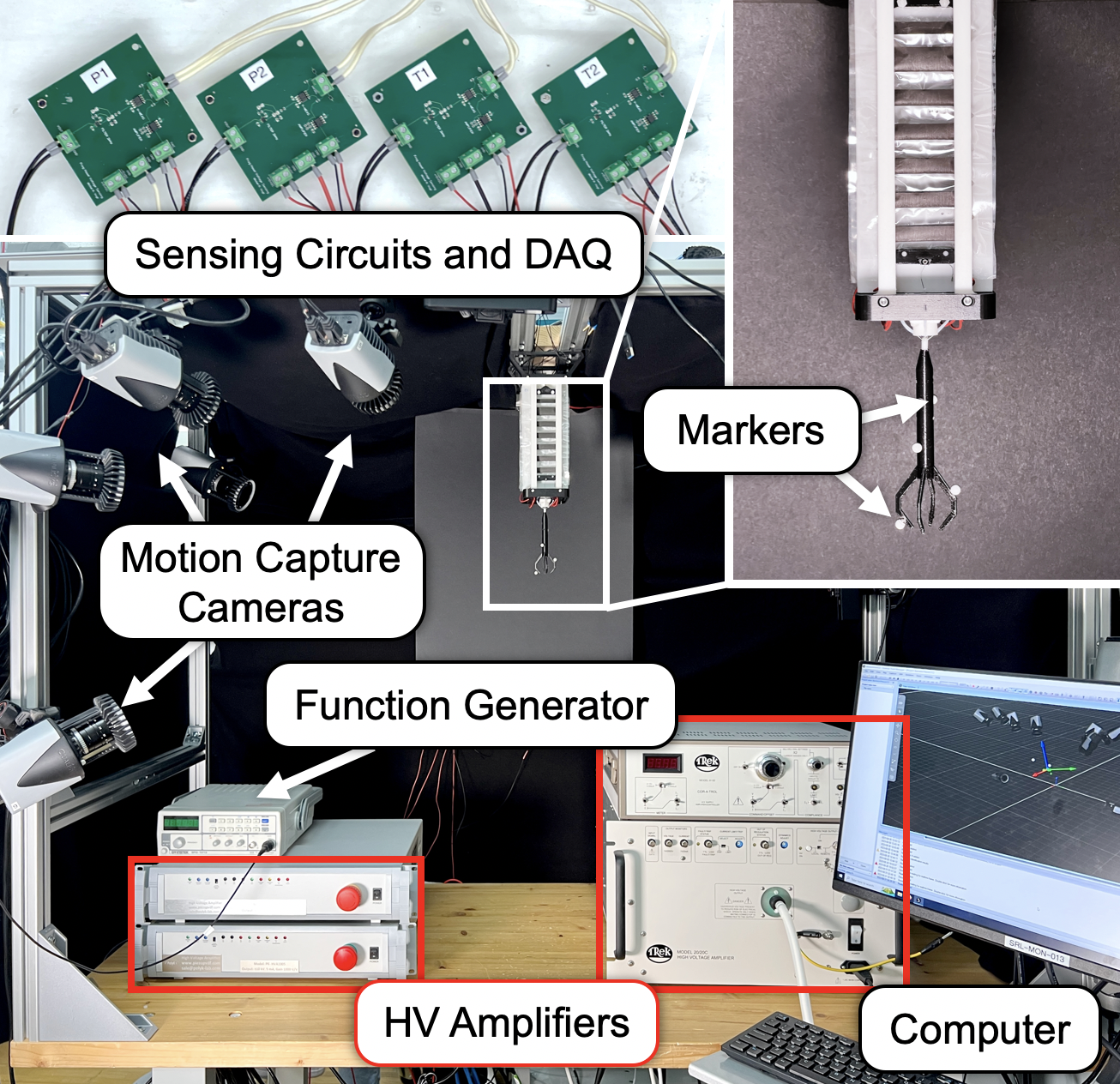}
\caption{Experimental Setup Overview: Four high-voltage amplifiers were placed inside the motion capture mobile rack. Low-voltage components, including the DAQ and sensing circuits, were located on top of the cage, separated from the high-voltage parts.}
\label{fig:exp-setup}
\vspace{-20pt}
\end{figure}

\Cref{fig:exp-setup} illustrates our experimental setup, comprising the following components:
\begin{itemize}
    \item A musculoskeletal robotic shoulder, as outlined in section \ref{system-design} with a gripper attached to the output link.
    \item A Qualisys motion capture system: Nine cameras tracked the six DoF of the manipulator in real-time at a sampling rate of \SI{150}{Hz}. To detect the manipulator's motion, four markers were attached to the robot's arm.
    \item Four high-voltage amplifiers supplied the required voltage range \numrange[range-phrase={--}]{0}{5.5} \si{\kilo\volt} for the HASEL actuators. The amplifiers used include one Trek 20/20C with a gain of 2000V/V, one Trek 610E, and two PolyK PK-HVA10005, each with a gain of 1000V/V. 
    \item A GW-INSTEK SGF-1013 generated the low voltage self-sensing signal for all HASEL actuators, outputting a \SI{10}{\volt} peak-to-peak with \SI{2}{\kHz}  sinusoidal signal.
    \item Two NI myDAQ devices captured the self-sensing voltages and outputted the commanded voltages for the high-voltage amplifiers. 
    \item Four self-sensing circuits measured the voltage drop across the HASELs and passed them to the DAQs.
    \item A custom \texttt{C++} application utilized the \texttt{myDAQmx} library to operate the DAQs at a sampling rate of \SI{200}{kHz} and the \texttt{RTClientSDK} library to stream data from the motion capture system. For data processing, a moving average filter with a window size of 20 was applied to the RMS self-sensing voltages. The application also estimated position and computed control signals within a control loop operating at \SI{200}{Hz}.
\end{itemize}
\subsection{Experiment Design}
In the design of our experiment, we began by planning trajectories in the task space. Using motion capture feedback, we adjusted the control parameters by increasing the P gain to just before instability, then raising the I gain until the steady-state error disappeared, and finally fine-tuning all PID gains for enhanced response times. This step was followed by training the estimation model, using the data logged during motion capture feedback control. Subsequently, the task space trajectories were tracked using self-sensing feedback control, during which we further fine-tuned the control parameters for improved performance.
\subsection{Task Space Control}
\begin{figure*}[t]
    \centering
    \includegraphics[width=\textwidth]{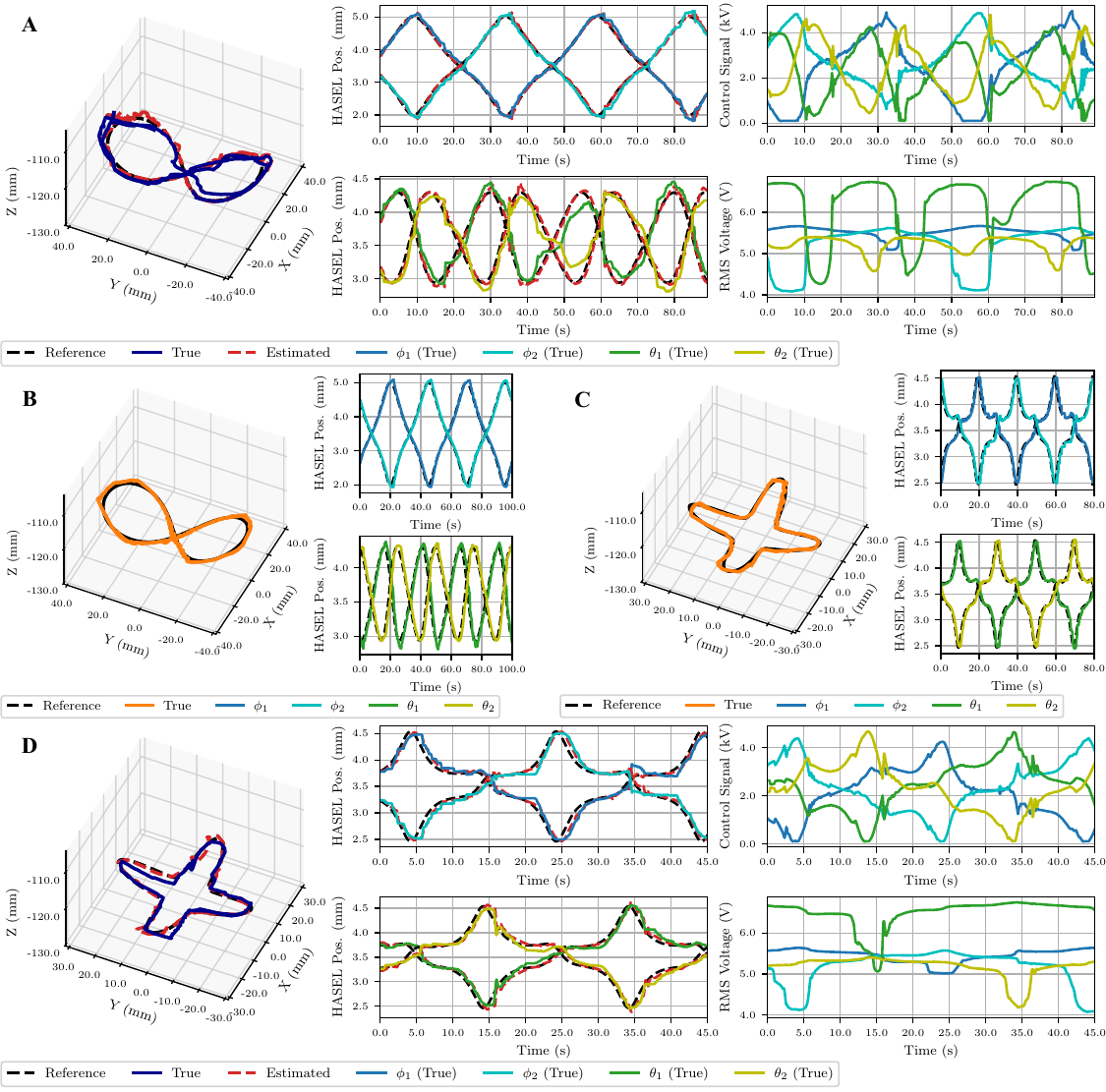}
    \caption{(A) Self-sensing control of a lemniscate trajectory with a period of \(T = 25s\). The top right plot represents the control signal of the four HASELs while the bottom right shows the RMS voltages of the self-sensing voltage drop. The central plots depict the reference, estimation, and true position in the tendon space. (B) Benchmark for the lemniscate trajectory: Closed-loop control using feedback position data from a motion capture system. The right plots show the reference and true HASEL position in the tendon space. (C) Benchmark for the star trajectory. This benchmark aligns with the format described in (B). (D) Self-sensing control of a star trajectory in the task space with a period of \(T = 40s\). The plot aligns with the format described in (A).}
    \label{fig:results-plot.png}
    \vspace{-22pt}
\end{figure*}

\Cref{fig:results-plot.png}
shows a qualitative comparison of the self-sensing feedback controller (SS) performance against the benchmark with motion capture feedback (BM). \Cref{tab:controller_data} shows the tuned PID gains for the conducted experiments. 

One issue with self-sensing control was the noisy estimated position, even after bandpass and digital filtering, which led to destabilization of the PID controller. We switched to a PI controller with lower P gains for self-sensing and implemented digital low-pass filters for both control and sensing. This strategy stabilized the system but introduced a slight phase lag, as shown in~\Cref{fig:results-plot.png}.

An analytical comparison for the lemniscate trajectory shows that the self-sensing control achieved an RMSE of \SI{4.245}{mm}, with the benchmark at \SI{2.869}{mm}. This indicates a \SI{48.0}{\percent} increase in error relative to the benchmark.
For the star trajectory, the self-sensing control RMSE was \SI{3.407}{mm} compared to the benchmark's \SI{2.798}{mm}. This translates to a \SI{21.8}{\percent} increase in error, suggesting a closer performance to the benchmark than observed in the lemniscate trajectory. 

\begin{table}[!h]
    \centering
    \caption{Controller Gains and RMSE Values. The amplifier for \(H_{\phi,1}\) has twice the output ratio, leading to notably smaller gains. Lemni stands for a lemniscate trajectory and Star stands for star trajectory.}
    \label{tab:controller_data}
    \begin{tabular}{lccc}
    \toprule
     \textbf{Trajectory} & \( \mathbf{K_p} \) & \( \mathbf{K_i} \times 10^{-3} \) & \( \mathbf{K_d} \) \\
    \midrule
    Lemni (SS) & [0.3, 0.75, 0.95, 0.85] & [1, 2, 2, 2] & [0, 0, 0, 0] \\
    Lemni (BM) & [0.4, 0.85, 1.05, 0.95] & [2, 3, 3, 3] & [0.5, 1, 1, 1] \\
    \midrule
    Star (SS) & [0.45, 0.85, 0.8, 0.90] & [1, 2, 2, 2] & [0, 0, 0, 0]  \\
    Star (BM) & [0.45, 0.95, 0.9, 0.95] & [1, 2, 2, 2] & [0.5, 1, 1, 1] \\
    \bottomrule
    \end{tabular}
\end{table}

\section{Conclusion}
In this paper, we introduced a bio-inspired antagonistic robotic shoulder with two DoF harnessing self-sensing Peano-HASEL actuators. This design eliminates external sensors, leading to a more compact and compliant mechanism. Our results underscore a step forward in the deployment of Peano-HASEL-driven robotic manipulators with inherent self-sensing capabilities, bridging the gap between mechanical and biological shoulder structures. 

Nevertheless, our study has limitations that suggest directions for future research. 
The current system faces a torque limitation due to the HASEL actuators' constraints, which could be mitigated by adding more actuators in parallel.
The estimation model serves as a cornerstone of our approach. While the model has been effective for the trajectories it was trained on, there is room for improvement when encountering new, unseen data. In future work, this improvement can be achieved by incorporating advanced machine learning techniques. On the topic of controllers, introducing a model-based controller could improve the performance and facilitate control at higher frequencies. Additionally, exploring the integration of a third yaw axis in actuation could lead to innovative control strategies, evolving towards a comprehensive three degrees of freedom actuation mechanism.

\begingroup
\footnotesize
\printbibliography
\endgroup
\end{document}